\documentclass[letterpaper]{article} 
\usepackage{aaai23}  
\usepackage{times}  
\usepackage{helvet}  
\usepackage{courier}  
\usepackage[hyphens]{url}  
\usepackage{graphicx} 
\usepackage{natbib}  
\usepackage{caption} 
\usepackage{algorithm}
\usepackage{algorithmic}
\usepackage{multirow}
\usepackage{amsmath,amsthm,amssymb,amsfonts}

\usepackage{newfloat}
\usepackage{listings}
\DeclareCaptionStyle{ruled}{labelfont=normalfont,labelsep=colon,strut=off} 
\floatstyle{ruled}
\newfloat{listing}{tb}{lst}{}
\floatname{listing}{Listing}
\title{EMEF: Ensemble Multi-Exposure Image Fusion}
\author{
    Renshuai Liu,
    Chengyang Li,
    Haitao Cao,
    Yinglin Zheng,
    Ming Zeng,
    Xuan Cheng\thanks{corresponding author}
}
\affiliations{

    School of Informatics, Xiamen University, Xiamen 361005, China
    
    \{medalwill, chengyanglee, chtao, zhengyinglin\}@stu.xmu.edu.cn,
    
    \{zengming, chengxuan\}@xmu.edu.cn
    
}

\usepackage{bibentry}

\begin{document}

\maketitle

\begin{abstract}
Although remarkable progress has been made in recent years, current multi-exposure image fusion (MEF) research is still bounded by the lack of real ground truth, objective evaluation function, and robust fusion strategy. In this paper, we study the MEF problem from a new perspective. We don’t utilize any synthesized ground truth, design any loss function, or develop any fusion strategy. Our proposed method EMEF takes advantage of the wisdom of multiple imperfect MEF contributors including both conventional and deep learning-based methods. Specifically, EMEF consists of two main stages: pre-train an imitator network and tune the imitator in the runtime. In the first stage, we make a unified network imitate different MEF targets in a style modulation way. In the second stage, we tune the imitator network by optimizing the style code, in order to find an optimal fusion result for each input pair. In the experiment, we construct EMEF from four state-of-the-art MEF methods and then make comparisons with the individuals and several other competitive methods on the latest released MEF benchmark dataset. 
The promising experimental results demonstrate that our ensemble framework can “get the best of all worlds”. 
The code is available at \url{https://github.com/medalwill/EMEF}.
\end{abstract}

\section{Introduction}
Real-world scenes usually exhibit a high dynamic range (HDR) that may be in excess of 100,000: 1 between the brightest and darkest regions.
The pictures captured by digital image sensors, however, usually have a low dynamic range (LDR), suffering from over-exposure and under-exposure in some situations. An effective yet economical solution is MEF, which fuses several LDR images in different exposures into a single HDR image.
Nowadays, MEF has already been widely used in smartphones like Xiaomi, vivo and OPPO.

\begin{figure}[t]
\centering
\includegraphics[width=1.0\columnwidth]{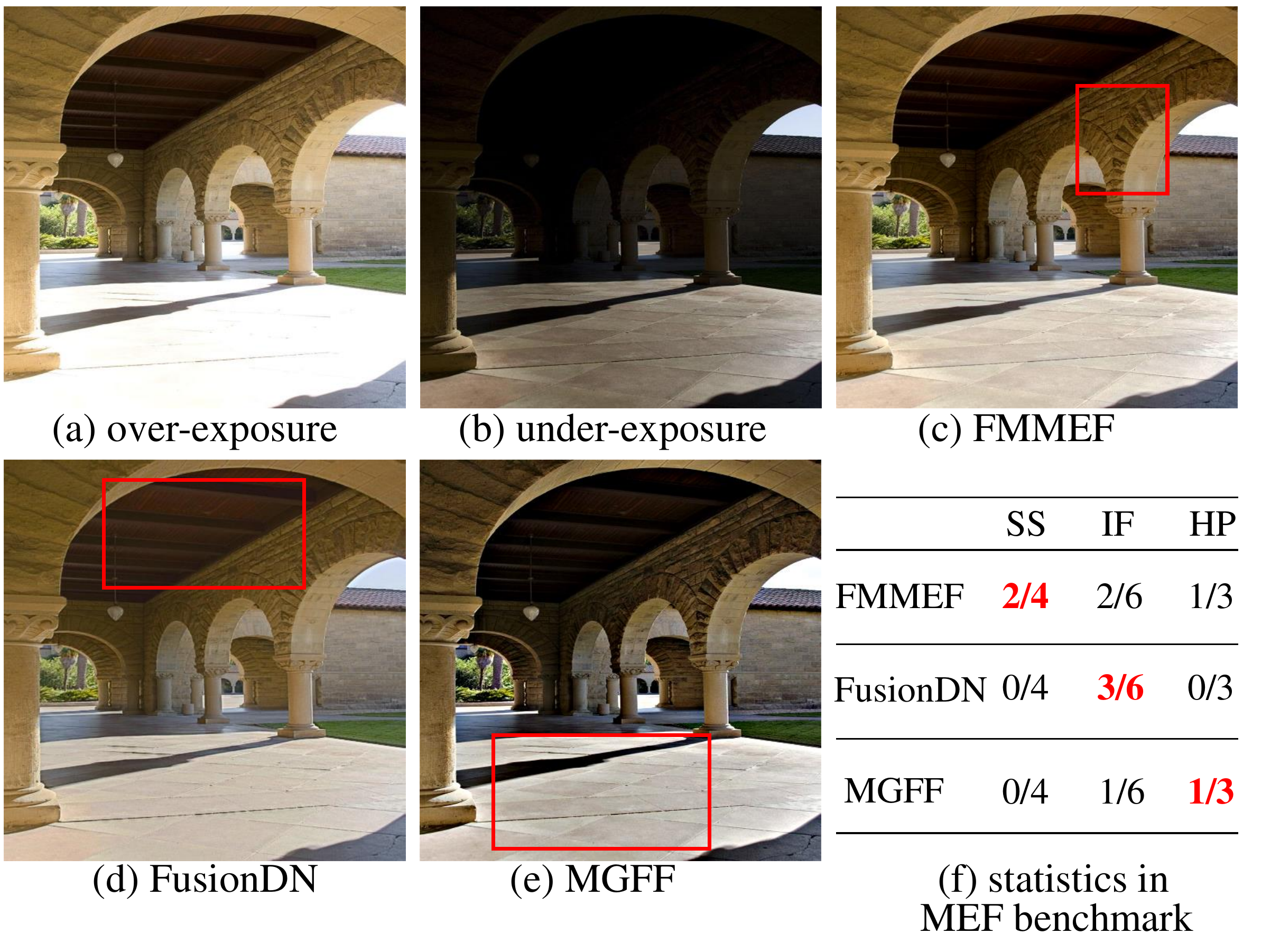} 
\caption{
The over-exposure (a) and under-exposure (b) images are the input. The fusion results of FMMEF (c), FusionDN (d), and MGFF (e) respectively exhibit more meaningful structures in the stone arch, roof, and ground regions. According to the evaluation (f) in MEFB \cite{zhang2021benchmarking}, FMMEF, FusionDN, and MGFF have advantages respectively in the metrics of structural similarity (SS), image feature (IF) and human perception (HP). ``2/4'' means that the method gets 2 best values among the 4 metrics.
}
\label{fig:banner}
\end{figure}

Many MEF methods have been proposed over the last decade. The traditional MEF methods use specific hand-crafted fusion strategies, while the deep learning-based MEF methods directly feed multi-exposure images into a network to produce a fused image in a supervised or unsupervised way. Although deep learning-based methods have gradually become mainstream in the MEF field, traditional methods still show very competitive performance in a recently published MEF benchmark (MEFB) \cite{zhang2021benchmarking}. Meanwhile, it's really hard to find a perfect MEF method at present, as no single method can always perform well in all situations. This is mainly due to three factors. 
1) The existing MEF ground truth data is mostly artificially made by selecting visually appealing results from a set of MEF methods. The lack of real ground truth hinders the ability of learning-based methods.
2) As HDR is a very subjective visual effect of human beings, there is no uniform objective metric that can well evaluate the fusion quality. Hence, the loss functions used in existing MEF methods are usually biased.
3) Most traditional methods make assumptions about the scenes, which are valid in some situations but invalid in others. It's really hard to design a one-size-fits-all image fusion strategy.

As a consequence, the existing MEF methods have their own strengths and weaknesses.
Based on the comprehensive quantitative evaluation of MEFB \cite{zhang2021benchmarking}, FMMEF \cite{FMMEF}, FusionDN \cite{FusionDN} and MGFF \cite{MGFF} are the top three methods.
FMMEF performs well in structural similarity-based metrics,
FusionDN exhibits good performance in image feature-based metrics, 
and MGFF gets high scores in human perception-inspired metrics. 
This phenomenon clearly dedicates that current state-of-the-arts have unique advantages when examined from different aspects.
We show an example in Fig. \ref{fig:banner}.

In this paper, we study the MEF problem from a fire-new perspective.
We don't utilize any synthesized ground truth, design any loss function, or develop any fusion strategy, like other traditional or deep learning-based MEF methods. 
Our proposed method takes advantage of the wisdom of multiple imperfect MEF methods, by combining each method's solution for the problem to give a higher quality solution than any individual.
We refer to our method as \emph{ensemble-based MEF} (EMEF),
as it shares a similar motivation with other ensemble methods \cite{ensembletracking2, ensembletracking} that combine multiple models.

The main contribution of this paper is the ensemble framework for the MEF problem. To realize the framework, we also propose several new network designs: 1) the imitator network which imitates different MEF methods' fusion effect in the unified GAN framework; 2) the optimization algorithm which searches the optimal code in the style space of the imitator to make MEF inference; 3) the random soft label to represent the style code, which removes artifacts while improves generalization.

\section{Related Work}
\label{related}
\subsection{Traditional MEF Methods}
Traditional MEF methods generally consist of spatial domain-based methods and transform domain-based methods. 

Spatial domain-based methods can be further divided into three categories, i.e., pixel-based, patch-based, and optimization-based methods. 
Pixel-based methods work on the pixel level, calculating the weighted sum of source images to derive a fused image. DSIFT-EF \cite{LIU2015208} estimates the weight maps of source images according to their local contrast, exposure quality, and spatial consistency, 
then refines the weight maps by a recursive filter.  
MEFAW\cite{MEFAW} defines two adaptive weights based on the relative intensity and global gradient of each pixel. The final weight maps are worked out with a normalized multiplication operation.
Different from the pixel-wised methods, patch-wised methods work on the patch level. The method proposed by \cite{7351094} decomposes each patch into signal strength, signal structure, and mean intensity, then reconstructs patches with the above components, and finally blends them to generate a fused image.
Based on the above method, SPD-MEF \cite{7859418} makes use of the direction of the signal structure component to achieve ghost removal. The representative of optimization-based methods is MEFOpt \cite{8233158}. This method introduces an evaluation metric named MEF-SSIM$_c$ which has improved performance compared to the original MEF-SSIM metric. Then the gradient descent is used to search the space of all images for a fusion result with optimal performance on MEF-SSIM$_c$.

Transform domain-based methods firstly transform source images to a specific domain to get their implicit representations,
then fuse these representations, and finally convert the fusion results back to the spatial domain by an inverse transform.
The method proposed by \cite{378222} is one of the first transform domain-based MEF methods which uses a gradient pyramid transform to get pyramid representations.
The method proposed by \cite{4392748} estimates the weight maps of source images considering their contrast, saturation, and exposure followed by a Gaussian filter smoothing. It also adopts a Laplacian pyramid transform to get laplacian coefficients which are then weighted according to the weight maps. Based on the above method,  \cite{li1995multisensor} fuse source images in the wavelet domain after a wavelet transform.

Although traditional methods have made great progress, they still have some drawbacks. e.g., it's not an easy task to design an effective fusion algorithm and there is no one-size-fits-all fusion strategy.

\begin{figure*}
\centering
\includegraphics[width=0.78\textwidth]{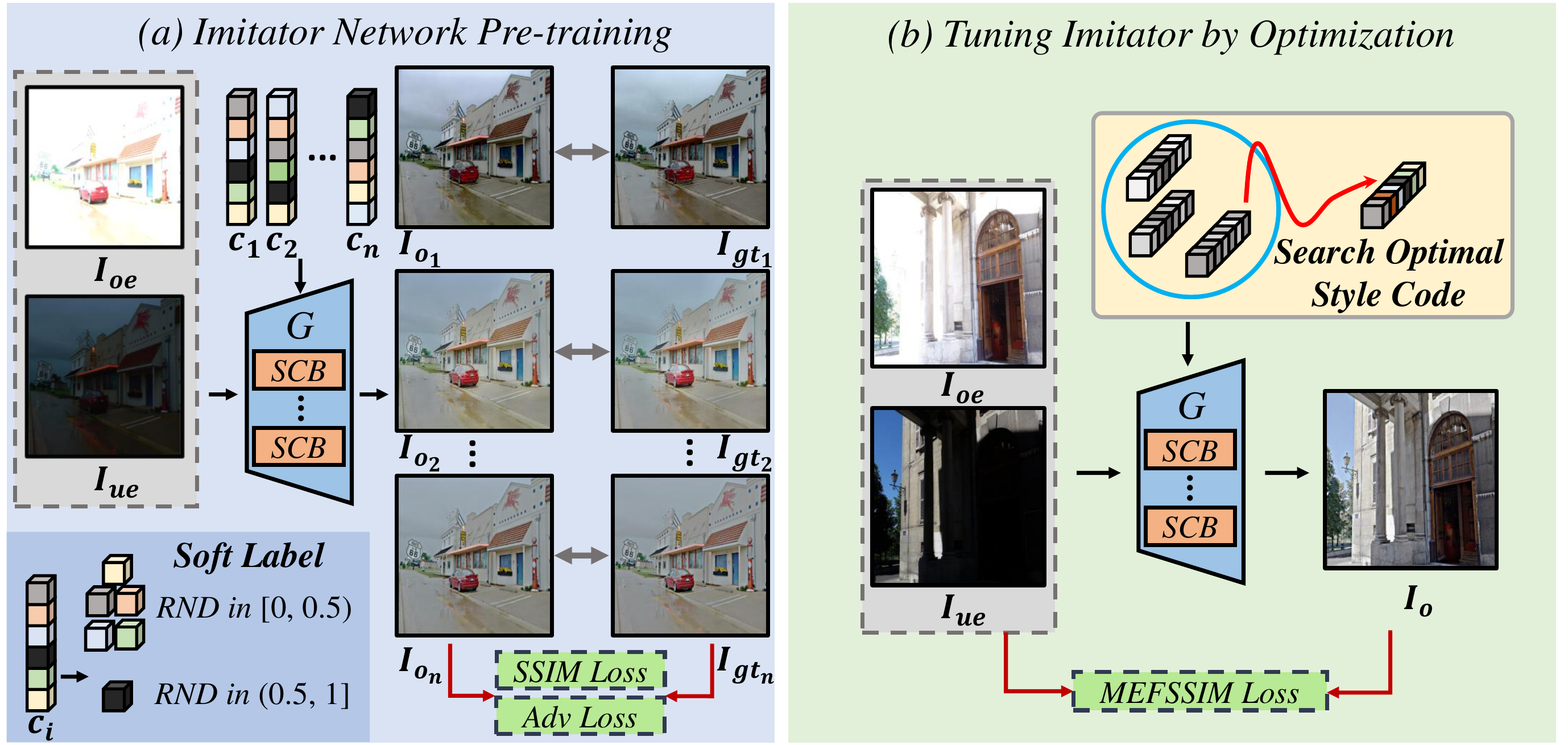} 
\caption{Overview. The proposed EMEF consists of two main stages: (a) pre-train an imitator network, and (b) tune the imitator in the runtime.}
\label{fig:pipeline}
\end{figure*}

\subsection{Deep Learning-Based MEF Methods}
Deep learning-based methods usually train networks in a supervised or unsupervised manner. As real ground truth data is hard to obtain, researchers attempt to synthesize ground truth by various means. EFCNN \cite{wang2018end} is one of the earliest supervised methods which makes ground truth by adjusting the pixel intensity of source images. SICE \cite{Cai2018deep} establishes a paired dataset by generating fusion results of existing methods and manually selecting the visually best one as ground truth. Based on the synthesized ground truth, MEF-GAN \cite{MEFGAN} makes progress by introducing GAN and self-attention mechanism into the field of MEF. CF-Net \cite{deng2021deep} suggests undertaking the super-resolution task and MEF task with a unified network so that collaboration and interaction between them can be achieved.

Another kind of method trains their networks in other tasks to learn image representations of source images, and then fuse these representations to reconstruct the final result. IFCNN \cite{zhang2020ifcnn} trains its model on a multi-focus fusion dataset. Similarly, transMEF \cite{qu2022transmef} applies three self-supervised image reconstruction tasks to capture the representations of source images.

Unsupervised methods take a different way to work without ground truth. They fuse the source images under the guidance of a specific image assessment metric. DeepFuse \cite{ram2017deepfuse} is not only the first unsupervised method but also the first deep-learning method, which works on YCrCb color space and applies MEF-SSIM to train its CNN. DIF-Net \cite{jung2020unsupervised} is designed for several image fusion tasks which focus on contrast preservation by employing a metric named structure tensor. U2Fusion \cite{xu2020u2fusion} measures the amount and quality of the information in source images by computing information preservation degree and adaptively fuses source images with respect to it. PMGI \cite{zhang2020rethinking} considers the fusion task as a proportional maintenance problem of gradient and intensity, which utilizes a two-branch network and divides the loss function into intensity and gradient parts.

\section{Method}
\subsection{Overview}
\label{over}
As shown in Fig. \ref{fig:pipeline}, the proposed EMEF consists of two main stages: pre-train an imitator network (Sect. \ref{inp}), and tune the imitator in the runtime (Sect. \ref{int}).
In the first stage, we utilize a unified network to imitate multiple MEF methods. Several traditional and deep learning-based MEF methods are selected as the imitation targets. We view each MEF target method as a ``style'' and then train a style-modulated GAN in a supervised way. 
Such a network can produce a very similar fusion result with each target method in the ensemble, under the control of a style code.
The style code determines which target methods the network would imitate in the online inference, and is represented by the random soft label (Sect. \ref{sl}) to improve the generalization ability.
In the second stage, we tune the pre-trained imitator network by searching the optimal style code, in order to make inferences for each input pair. 
An image quality assessment-based loss function is optimized in the gradient descent way. Finally, EMEF is able to produce the best fusion result from the combined space of the MEF target methods.

\subsection{Imitator Network Pre-training}
\label{inp}
Before constructing the imitator network, we collect the training data for it. For a pair of over-exposed and under-exposed images denoted by $I_{oe}, I_{ue}$, we use all the MEF target methods $\mathcal{M}_i, i=1,2,..n$ in the ensemble to produce their fusion images $I_{gt_i}$. Besides, each $\mathcal{M}_i$ is represented by a certain style code $c_i \in \mathbb{R}^n$. 
A sample in the training data can be formulated as:
\begin{equation}
\label{eq:trainingdata}
[I_{oe}, I_{ue},\{I_{gt_i}, c_i\}_{i=1,2,..n}].
\end{equation}
We believe that the particular fusion strategy and loss functions used in $\mathcal{M}_i$ have already been embedded in our constructed training data, and can be learned by the deep models.

The core of the imitator network is a style-modulated generator denoted by $\mathcal{G}$. As shown in Fig. \ref{fig:pipeline}, the generator takes the image pair $I_{oe}$, $I_{ue}$, and the style code
$c_i$ as input, and outputs the fusion result $I_{o_i}$. Such generation process can be formulated as:
\begin{equation}
\label{eq:generator}
I_{o_i} = \mathcal{G}(I_{oe}, I_{ue}, c_i, \theta),
\end{equation}
where $\theta$ is the parameters of $\mathcal{G}$.
We require $I_{o_i}$ to match its corresponding $I_{gt_i}$ as much as possible and thus train $\mathcal{G}$ in a supervised manner.  

\begin{figure*}
\centering
\includegraphics[width=0.8\textwidth]{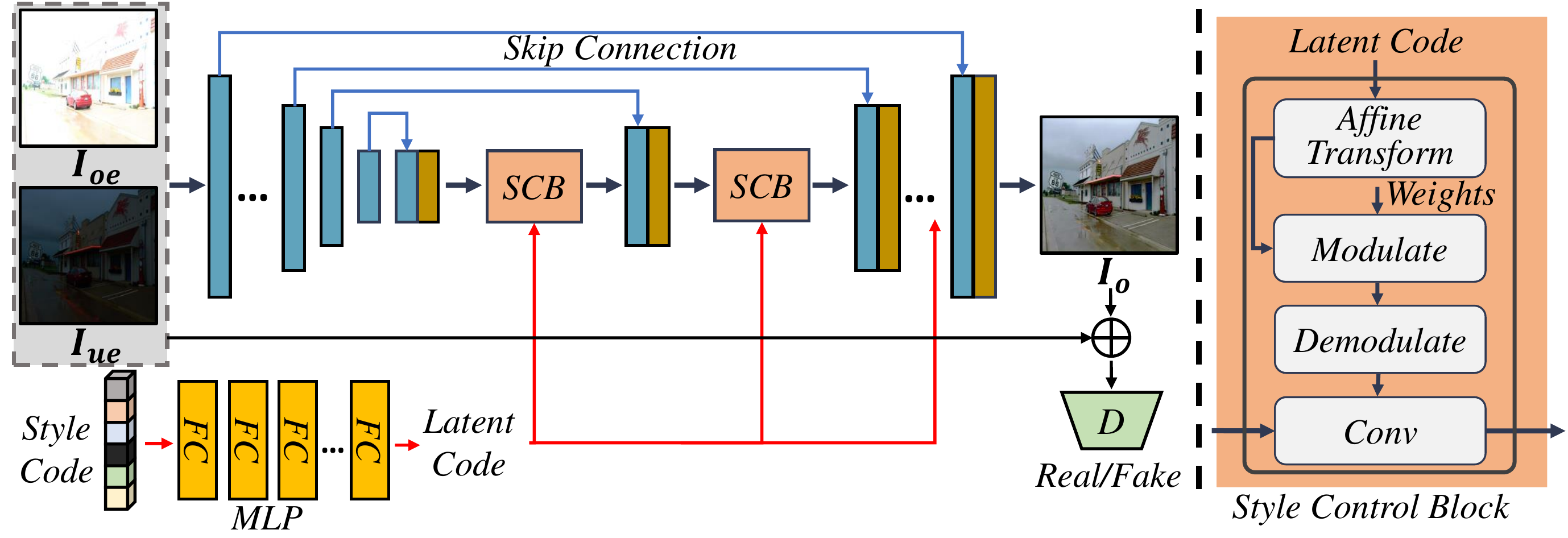} 
\caption{Generator. The network architecture consists of the UNet, several Style Control Blocks (SCB), and the MLP-based mapping block.}
\label{fig:generator}
\end{figure*}

The network architecture of $\mathcal{G}$ is shown in Fig. \ref{fig:generator},
which adopts a standard UNet as the backbone and incorporates the Style Control Block (SCB).
The UNet extracts multi-scale features from the input images in the encoder and adds them back to each layer in the decoder, which helps to preserve more information from the input. The SCB injects the style code $c_i$ to each layer in the decoder of UNet except the last one, which is the key to our style control. The style code is not directly used but mapped into a latent space by a Multilayer Perceptron (MLP) before being fed to SCB.

\textbf{Style Control Block.} 
We leverage the merit of StyleGAN2 \cite{karras2020analyzing} to construct SCB.
SCB consists of a convolution layer and two operations (modulation and demodulation) to its weights. 
For an input latent code $l$, SCB firstly transforms the $l_i$ of the $i$th layer 
into the micro style $s_i$ with an affine transformation, so that the latent code $l$ can match the scale of different layers.
Then the weights are scaled with the $s_i$ to fuse $s_i$ into the activation, which helps to decouple styles of different target methods. The weight modulation operation can be formulated as:
\begin{equation}
\label{eq:mod}
w_{ijkl}^{'} = s_j \cdot w_{ijkl},
\end{equation}
where $w$ denotes the original weights, $w'$ denotes the modulated weights, $i$ denotes the $i$-th output channel of the weight, $j$ denotes the $j$-th input channel of the weight,
and $(k,l)$ denotes the coordinate of the convolution kernel.
Subsequently, weight demodulation is conducted, which shrinks the weights to keep the statistics of activations unchanged. It's formulated as:
\begin{equation}
\label{eq:demod}
w_{ijkl}^{''} = w_{ijkl}^{'}\bigg/ \sqrt{\sum_{j,k,l} {w_{ijkl}^{'}}^2 + \epsilon}
\end{equation}
where $\epsilon$ is a small positive constant to promote robustness.

\begin{figure}
\centering
\includegraphics[width=0.47\textwidth]{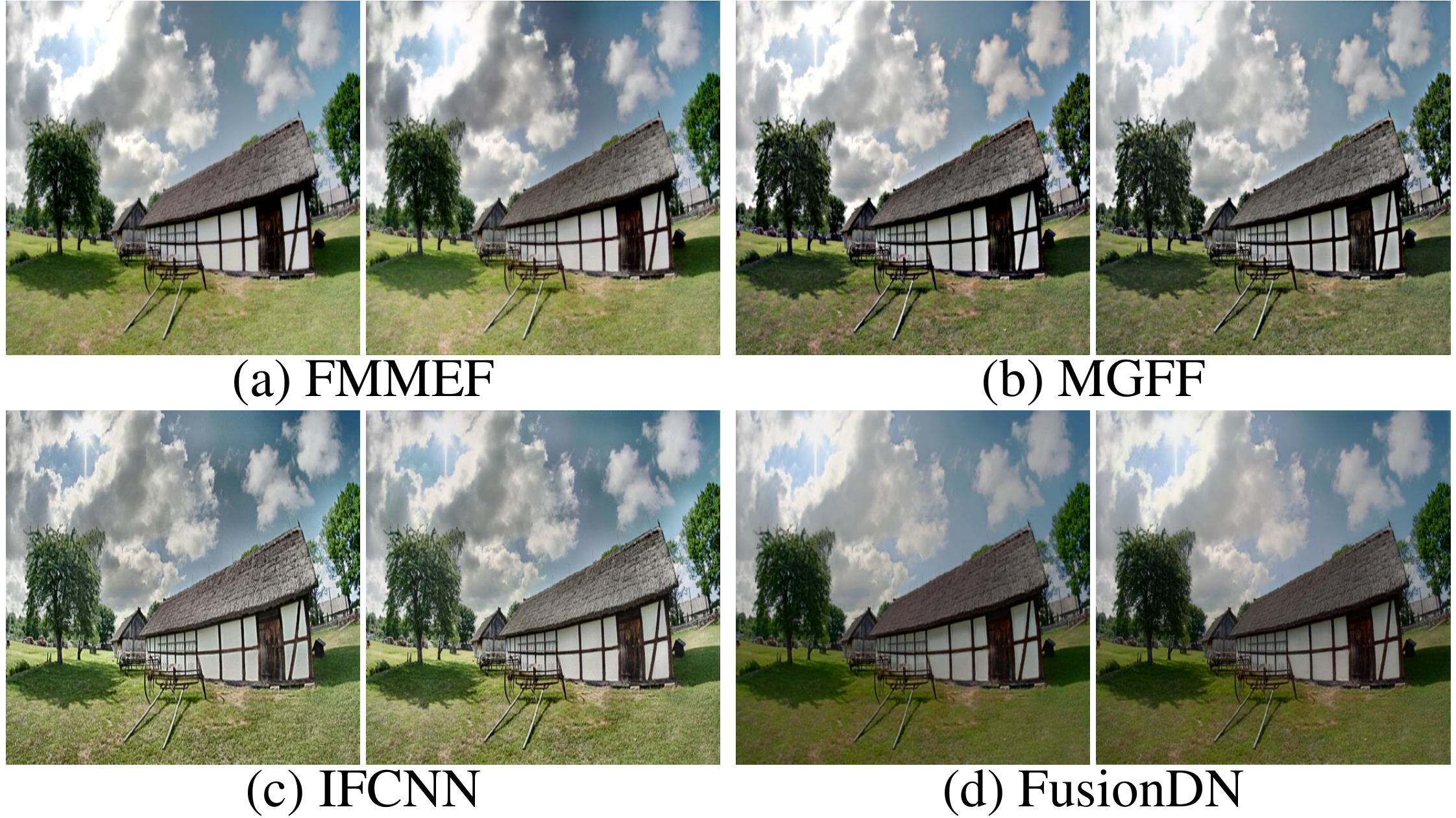} 
\caption{An example of our MEF imitation. In each pair, the left one is the fusion result $I_{gt_i}$ from the imitation target method while the right one is the imitative result $I_{o_i}$ from our imitator network.}
\label{fig:ablation_imitation}
\end{figure}

\textbf{Loss.} 
The imitator network is optimized by minimizing SSIM loss and adversarial loss.
SSIM loss measures the structural similarity between $I_{o_i}$ and its corresponding ground truth $I_{gt_i}$, which can be formulated as:
\begin{equation}
\label{eq:SSIMLoss}
L_{SSIM} = 1 - SSIM(I_{o_i}, I_{gt_i}).
\end{equation}
$SSIM(\cdot, \cdot)$ denotes the standard SSIM metric.
To promote the realism of $I_{oi}$, we also employ adversarial loss.
It's formulated as:
\begin{equation}
\label{eq:GANLoss}
\begin{split}
L_{adv} = \mathbb{E}(1-\log \mathcal{D}(I_{o_i}, I_{oe}, I_{ue}, \gamma)) \\ 
+\mathbb{E}(\log \mathcal{D}(I_{gt_i}, I_{oe}, I_{ue}, \gamma)),
\end{split}
\end{equation}
where $\mathcal{D}$ is the discriminator and $\gamma$ is its parameters.

The final loss is the weighted sum of the aforementioned losses:
\begin{equation}
\label{eq:total}
\begin{split}
L_{pre} = L_{SSIM} + \lambda L_{adv}
\end{split}
\end{equation}
where $\lambda$ is a trade-off between the two losses.
Fig. \ref{fig:ablation_imitation} shows our imitation results.
There is little visual difference between the output $I_{o_i}$ of our imitator network and the fusion result of the target method $I_{gt_i}$.

\subsection{Imitator Network Tuning}
\label{int}

\begin{algorithm}
\caption{Search for the optimal style code $c^* \in \mathbb{R}^n$}
\label{alg1}
\begin{algorithmic}[1]
\INPUT
    A pair of over-exposed and under-exposed images $I_{oe}, I_{ue}$.
    The pre-trained imitator network $\mathcal{G}$.
\INITIALIZE
    Initialize the style code $c^*$ with an all-one vector.  
    Concatenate $I_{oe}, I_{ue}$ into $I_{oue}$ in the channel dimension.
\REPEAT
\STATE $L \leftarrow 1 - MEFSSIM(I_{oue}, \mathcal{G}(I_{oe}, I_{ue}, c^*))$
\STATE $c^* \leftarrow c^* - \alpha \nabla_{c^*} L$
\UNTIL $converged$
\OUTPUT $c^*$
\end{algorithmic}
\end{algorithm}

In this stage, we tune the style code in the pre-trained imitator network to make inferences for an input pair. As we mentioned before, there is no perfect MEF method. 
It's better to utilize different suitable MEF methods for different types of source images.
To realize the goal, we search for an optimal style code for the input pair.
The pseudo-code of the searching procedure is presented in Algorithm \ref{alg1}.
Starting from an all-one initialization, we use the gradient descent algorithm to search for an optimal style code $c^*$ that minimizes the MEF-SSIM \cite{MEF-SSIM} loss function. The MEF-SSIM measures how much vital information from input images can be preserved in the fused image, and is a frequently used MEF image quality assessment model.

\subsection{Random Soft Label}
\label{sl}
\begin{figure}
\centering
\includegraphics[width=0.47\textwidth]{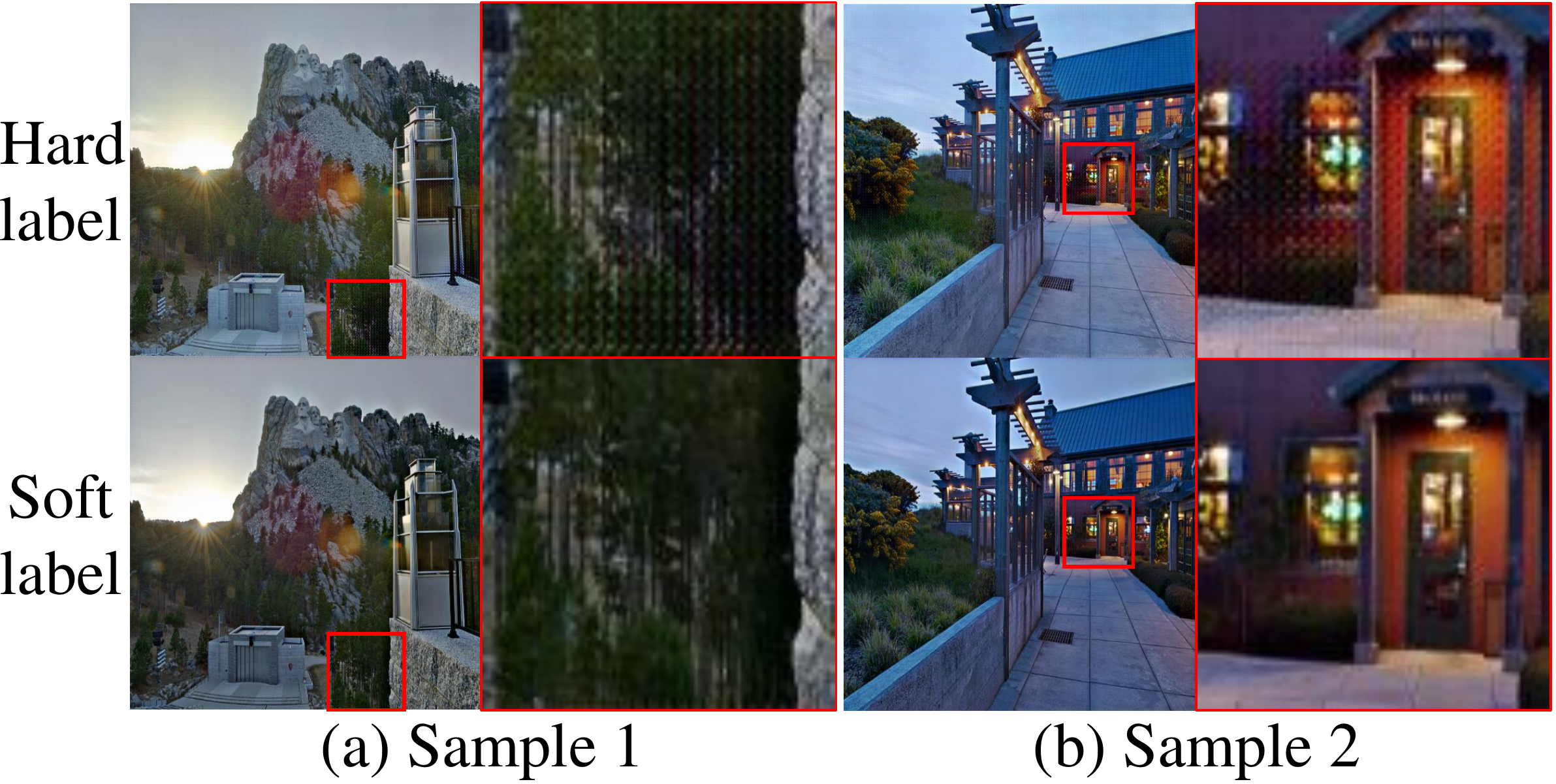} 
\caption{Hard label introduces unacceptable image artifacts, such as stripes and grids, while random soft label can remove them by mitigating the domain gap.}
\label{fig:softlabel}
\end{figure}

Intuitively, we use the one-hot label (e.g. $\{0, 1, 0, 0\}$) as the style code in the imitator network pre-training. However, the optimized style codes obtained in the imitator tuning stage are usually floats (e.g. $\{0.22, 0.15, 0.85, 0.36\}$) rather than integers. The significant domain gap between the style codes in pre-training and tuning introduces severe artifacts. We show an example in Fig. \ref{fig:softlabel}.
To overcome this issue, we adopt a random soft label trick that can mitigate the domain gap and promote robustness. We replace the 1 value in the one-hot label with a random number in the range of $(0.5, 1.0]$, and replace the 0 value with a random number in the range of $[0.0, 0.5)$. For example, $\{0, 1, 0, 0\}$ is replaced by $\{0.08, 0.73, 0.36, 0.21\}$. The experimental results demonstrate that the random soft label 
eliminates artifacts, mitigates the domain gap and greatly 
improves the generalization ability of the network.

\subsection{Discussions}
\textbf{Relationship with supervised MEF.}
The supervised MEF methods \cite{wang2018end, Cai2018deep, MEFGAN} directly optimize the reconstruction loss between the fusion result and the ground truth. However, the ground truth is usually artificially made. Our proposed EMEF uses the supervised way only for imitating different MEF targets in a unified framework, but not for generating the fusion result.

\textbf{Relationship with unsupervised MEF.}
The unsupervised MEF methods \cite{ram2017deepfuse, xu2020u2fusion} usually optimize the loss that measures the retention degrees of image features from the input. We also use such unsupervised loss in the MEF inference. The unsupervised method searches the entire image space, while our proposed EMEF takes the pre-trained imitator network as the prior, thus constructing a smaller space (the combined space of the MEF target methods). 
Hence, EMEF searches for a low-dimension style code rather than a high-dimension image, which provides the one-shot MEF and increases robustness.  

\textbf{Relationship with ensemble GAN.}
There are ensemble methods \cite{Arora2017,MGAN2018,GANensemble2021} that train GAN with multiple generators rather than a single one, thus delivering a more stable image generation. Our method uses a unified generator to model different data distributions of the fusion results in different MEF methods. Another difference is that the ensemble GANs usually produce the final result by averaging or randomly selecting the generators' output, while our method finds an optimal result by optimization in the generator's style space.

\section{Experiments}
\subsection{Implementation Details}
In our experiments, $\lambda$ is set to 0.002. The network architecture of the generator follows the image-to-image translation network \cite{cite_pix2pix}. The image size of both input and output are 512 $\times$ 512. In the imitator network pre-training, the batch size is set to 1 and the network is trained with an Adam optimizer for 100 epochs. 
In the first 50 epochs, the learning rate is set to $2e-4$, and then decays linearly for the rest. 
In the imitator tuning, we adopt an adaptive search strategy with a 
20-step linear learning rate decay.
We choose the top-four MEF methods in MEFB to construct the ensemble, including FMMEF, FusionDN, MGFF, and IFCNN, and implement EMEF with Pytorch.
All experiments are conducted with two GeForce RTX 3090 GPUs.
It takes about 1.8 minutes to generate a $512\times512$ fusion image.

\subsection{Experimental Settings}
\textbf{Datasets.}
We train EMEF with the SICE \cite{Cai2018deep} dataset and evaluate it in MEFB \cite{zhang2021benchmarking}. SICE contains 589 image sequences of different exposures, and each sequence has a selected fused image attached as ground truth. We focus on the extreme static MEF problem so that only the brightest image and the darkest image within 356 static scene sequences are selected as the training data. MEFB contains 100 over-exposure and under-exposure image pairs captured under various conditions, which can provide fair and comprehensive comparisons.

\textbf{Evaluation Metrics.}
We apply 12 metrics from four perspectives to evaluate the proposed EMEF.
The metrics include information theory-based metrics,
$\rm CE$ \cite{2009Image}, $\rm EN$ \cite{2008Assessment},
$\rm PSNR$ \cite{2015A}, $\rm TE$ \cite{2006Image};
image feature-based metrics,
$\rm AG$ \cite{2015Detail}, $\rm EI$ \cite{2018Hybrid},
$\rm Q^{AB/F}$ \cite{2000Objective}, 
$\rm Q_P$ \cite{zhao2007performance}, $\rm SF$ \cite{1995Image};
structural similarity-based metrics,
$\rm Q_W$ \cite{piella2003new},
$\rm MEF-SSIM$ \cite{MEF-SSIM};
human perception inspired metrics, $\rm Q_{CV}$ \cite{chen2007human}.

\begin{figure*}
\centering
\includegraphics[width=0.95\textwidth]{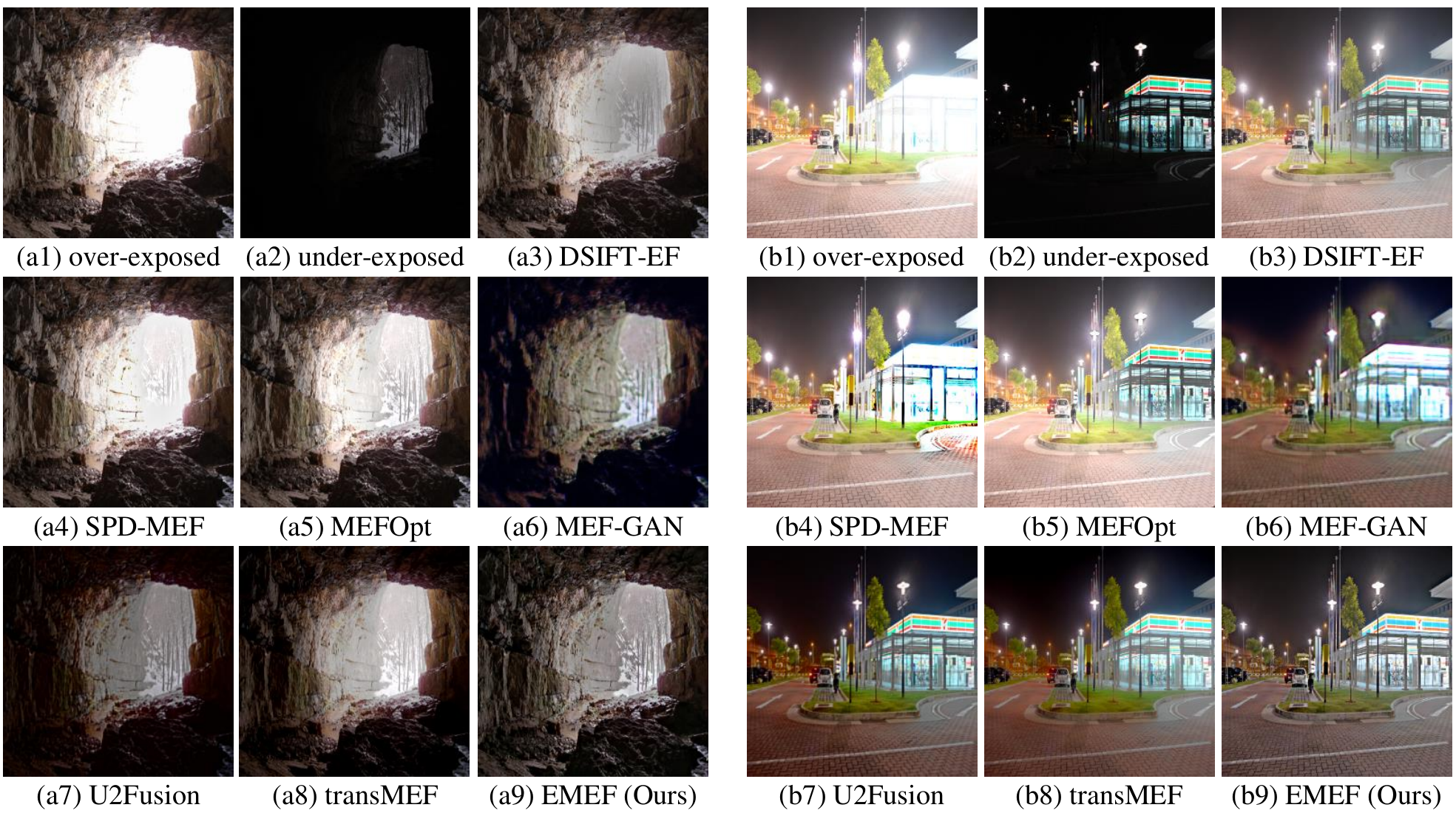} 
\caption{Qualitative comparison of EMEF with 6 competitive MEF methods on 2 typical multi-exposure image pairs in the MEFB dataset.}
\label{fig:qualitative}
\end{figure*}

\begin{table*}
\centering
\setlength\tabcolsep{1pt}
\renewcommand\arraystretch{1.1}
\resizebox{\linewidth}{!}{
\begin{tabular}{c|cccccccccccc|c}
\hline
Methods            & $\rm CE\downarrow$ & $\rm EN\uparrow$ & $\rm PSNR\uparrow$ & $\rm TE\uparrow$ & $\rm AG\uparrow$ & $\rm EI\uparrow$ & $\rm Q^{AB/F}\uparrow$ & $\rm Q_P\uparrow$ & $\rm SF\uparrow$ & $\rm Q_W\uparrow$ & \begin{tabular}[c]{@{}c@{}}$\rm MEF$-\\ $\rm SSIM\uparrow$\end{tabular} & $\rm Q_{CV}\downarrow$ & \begin{tabular}[c]{@{}c@{}}Overall\\ Rank$\downarrow$\end{tabular} \\ \hline
DSIFT-EF           & 1.3026(1)         & 7.261(1)       & 53.336(6)        & 9866.552(3)    & 5.836(7)       & 59.269(7)      & 0.740(1)             & 0.769(1)        & 18.836(6)      & 0.865(3)        & 0.855(3)                   & 519.914(6)           & 45                                                                 \\
SPD-MEF            & 2.6909(6)         & 7.112(4)       & 53.594(3)        & 8801.729(5)    & 6.890(3)       & 69.691(3)      & 0.683(4)             & 0.737(2)        & 23.086(1)      & 0.822(6)        & 0.833(5)                   & 383.757(4)           & 46                                                                 \\
MEFOpt             & 2.3232(5)         & 7.195(3)       & 53.231(7)        & 11489.361(2)   & 6.924(2)       & 69.939(2)      & 0.731(2)             & 0.698(5)        & 22.188(3)      & 0.899(1)        & 0.866(2)                   & 578.273(7)           & 41                                                                 \\
MEF-GAN            & 1.8621(3)         & 6.933(5)       & 53.520(5)        & 7130.901(7)    & 5.874(6)       & 62.075(5)      & 0.442(7)             & 0.388(7)        & 17.308(7)      & 0.611(7)        & 0.761(7)                   & 394.722(5)           & 71                                                                 \\
U2Fusion           & 1.9372(4)         & 6.609(7)       & 53.621(2)        & 20056.882(1)   & 5.986(4)       & 62.349(4)      & 0.596(6)             & 0.654(6)        & 19.021(5)      & 0.826(5)        & 0.849(4)                   & 279.322(2)           & 50                                                                 \\
transMEF           & 2.7162(7)         & 6.768(6)       & 53.581(4)        & 8393.075(6)    & 5.891(5)       & 59.780(6)      & 0.682(5)             & 0.722(4)        & 19.648(4)      & 0.836(4)        & 0.832(6)                   & 247.126(1)           & 58                                                                 \\
\textbf{EMEF} & 1.7607(2)         & 7.219(2)       & 53.624(1)        & 9134.513(4)    & 6.969(1)       & 70.177(1)      & 0.693(3)             & 0.725(3)        & 22.755(2)      & 0.885(2)        & 0.875(1)                   & 312.892(3)           & \textbf{25}                                                        \\ \hline
\end{tabular}
}
\caption{Quantitative results of EMEF and several MEF competitors (DSIFT-EF, SPD-MEF, MEFOpt, MEF-GAN, U2Fusion, transMEF) over MEFB in 256 × 256 resolution. All metrics except $\rm CE$ and $\rm Q_{cv}$ follow ``higher is better". The number listed within the bracket after each score denotes the rank in the metric. The overall rank is the sum of ranks on all metrics.}
\label{tab:main}
\end{table*}

\begin{table*}
\centering
\setlength\tabcolsep{0.6pt}
\renewcommand\arraystretch{1.1}
\resizebox{\linewidth}{!}{
\begin{tabular}{c|cccccccccccc|c}
\hline
Methods              & $\rm CE\downarrow$ & $\rm EN\uparrow$ & $\rm PSNR\uparrow$ & $\rm TE\uparrow$ & $\rm AG\uparrow$ & $\rm EI\uparrow$ & $\rm Q^{AB/F}\uparrow$ & $\rm Q_P\uparrow$ & $\rm SF\uparrow$ & $\rm Q_W\uparrow$ & \begin{tabular}[c]{@{}c@{}}$\rm MEF$-\\ $\rm SSIM\uparrow$\end{tabular} & $\rm Q_{CV}\downarrow$ & \begin{tabular}[c]{@{}c@{}}Overall\\ Rank$\downarrow$\end{tabular} \\ \hline
FMMEF                & 2.7538(6)         & 7.146(7)       & 56.557(5)        & 16371.098(8)   & 5.052(9)       & 51.808(9)      & 0.765(1)             & 0.760(1)        & 16.951(9)      & 0.914(1)        & 0.893(3)                   & 417.983(7)           & 66                                                                 \\
MGFF                 & 2.8829(9)         & 7.088(8)       & 56.603(2)        & 7914.649(9)    & 6.096(2)       & 62.546(2)      & 0.692(3)             & 0.740(2)        & 20.502(2)      & 0.860(4)        & 0.884(6)                   & 346.728(1)           & 50                                                                 \\
IFCNN                & 2.8488(8)         & 7.303(1)       & 56.422(7)        & 37365.314(3)   & 8.190(1)       & 80.417(1)      & 0.562(9)             & 0.618(6)        & 26.253(1)      & 0.791(8)        & 0.842(9)                   & 450.971(8)           & 62                                                                 \\
FusionDN             & 2.7072(5)         & 7.242(4)       & 56.397(8)        & 56725.167(2)   & 5.375(7)       & 55.559(7)      & 0.589(8)             & 0.588(7)        & 16.979(8)      & 0.789(9)        & 0.868(8)                   & 371.430(4)           & 77                                                                 \\
Pick $I_{gt}^*$      & 2.7665(7)         & 7.181(6)       & 56.535(6)        & 18988.347(7)   & 5.368(8)       & 55.086(8)      & 0.720(2)             & 0.721(3)        & 17.722(7)      & 0.883(2)        & 0.887(5)                   & 372.248(5)           & 66                                                                 \\
Pick $I_o^*$         & 1.7470(1)         & 7.271(3)       & 56.563(4)        & 19542.253(6)   & 5.590(4)       & 57.215(4)      & 0.640(5)             & 0.654(5)        & 18.661(4)      & 0.844(7)        & 0.889(4)                   & 400.537(6)           & 53                                                                 \\
opt. latent code & 2.1209(4)         & 6.774(9)       & 56.384(9)        & 81545.751(1)   & 5.565(5)       & 56.158(5)      & 0.610(7)             & 0.477(9)        & 17.934(6)      & 0.850(6)        & 0.955(1)                   & 714.371(9)           & 71                                                                 \\
w./o. soft label     & 1.9037(3)         & 7.272(2)       & 56.584(3)        & 28799.171(4)   & 6.054(3)       & 60.470(3)      & 0.613(6)             & 0.585(8)        & 19.542(3)      & 0.854(5)        & 0.873(7)                   & 370.834(3)           & 50                                                                 \\
\textbf{EMEF} & 1.9035(2)         & 7.239(5)       & 56.618(1)        & 22898.593(5)   & 5.504(6)       & 56.130(6)      & 0.667(4)             & 0.665(4)        & 18.319(5)      & 0.876(3)        & 0.898(2)                   & 365.065(2)           & \textbf{45}                                                        \\ \hline
\end{tabular}

}
\caption{Quantitative results of EMEF, the methods in the ensemble (FMMEF, MGFF, IFCNN, FusionDN), and the methods in ablation study (pick $I_{gt}^*$,
pick $I_{o}^*$, optimize latent code, w./o. soft label) over MEFB
in 512 × 512 resolution. }
\label{tab:ablation}
\end{table*}

\textbf{Competitors.}
Besides the four MEF methods included in the ensemble, we choose other 6 competitive MEF methods as competitors, which include three traditional methods, DSIFT-EF \cite{LIU2015208}, SPD-MEF \cite{7859418}, MEFOpt \cite{8233158}; and three deep learning-based methods, MEF-GAN \cite{MEFGAN}, U2Fusion \cite{xu2020u2fusion}, transMEF \cite{qu2022transmef}. 

\subsection{Comparisons with the MEF Competitors}

The qualitative comparison of EMEF with the 6 competitors is shown in Fig. \ref{fig:qualitative}. 
The main goal of MEF is to make the dark region brighter while making the bright region darker so that more details can be maintained. In sample (a), the region inside the cave is dark and the region outside is bright. SPD-MEF, MEFOpt failed to darken the bright region while MEF-GAN, U2Fusion was unable to brighten the dark region. DSIFT-EF and transMEF manage to behave well in both regions but exhibit lower contrast and fewer details outside the cave compared with our EMEF. In sample (b), the goal is to provide appropriate exposure for the streetlights, the store, and the floor. DSIFT-EF, SPD-MEF, MEFOpt, and MEF-GAN failed to achieve this goal, as terrible halo artifacts surround the streetlights and the store. Our method produces more appealing luminance in the left part of the fused image than U2Fusion. When compared with transMEF, our method generates far finer details in the store and the floor regions.

The quantitative comparison is presented in Table \ref{tab:main}.
The existing MEF methods are capable of achieving good performance on their preference metrics. Due to the extreme pursuit of these metrics, they usually show poor scores on the remaining metrics. The overall rank is the sum of ranks on all metrics which can reveal overall performance. We integrate 4 distinctive methods in EMEF so that they compensate each other. Thus our method has a relatively all-round ability on all metrics, ultimately presenting balanced and optimal overall performance.

\subsection{Comparisons with the MEF Methods in the Ensemble}
The qualitative comparison of EMEF with the 4 methods included in the ensemble is shown in Fig. \ref{fig:ablations}.
In sample (a), MGFF and FusionDN fail to brighten the dark regions in the sea, while FMMEF and IFCNN don't recover more details in the sky compared with ours.
In sample (b), MGFF fails to brighten the dark region in the grass,
FMMEF and IFCNN exhibit a little over-exposure in the grass,
and FusionDN generates an image of low contrast and poor luminance.
In both samples, our method can reconstruct the scene with moderate lighting conditions.
The quantitative comparison is presented in Table \ref{tab:ablation}.
Our method surpasses the methods included in the ensemble in the overall performance due to its integrated capability.

\subsection{Ablation Study}
\begin{figure}
\centering
\includegraphics[width=0.475\textwidth]{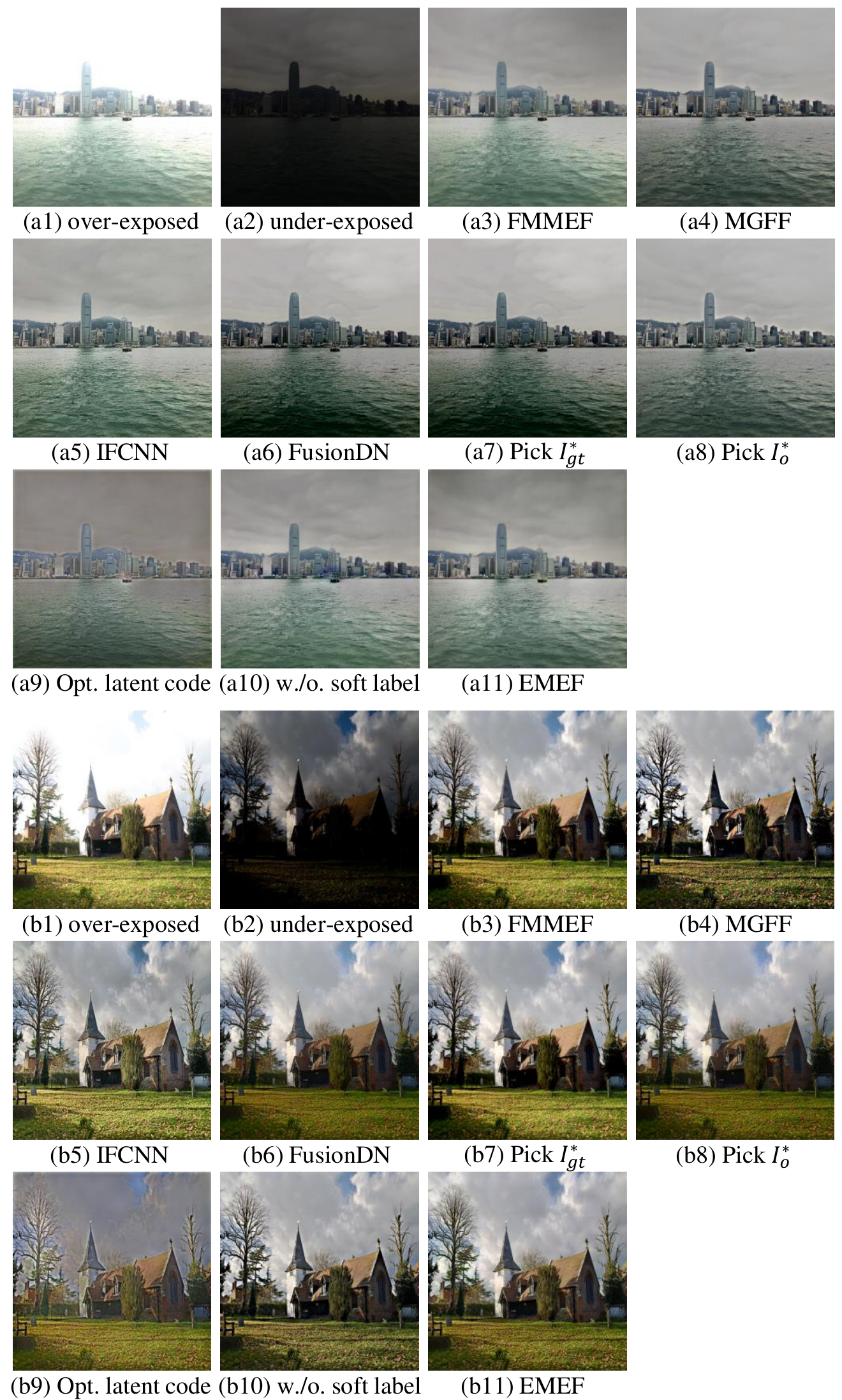} 
\caption{Qualitative comparison of EMEF with the four MEF methods in the ensemble and the four methods in the ablation study.}
\label{fig:ablations}
\end{figure}

In the ablation study, we evaluate other four methods: 1) selecting the MEF result $I_{gt}^*$ directly from the outputs of the four methods in the ensemble with the highest MEF-SSIM score, 2) selecting the MEF result $I_o^*$ directly from the four imitation results of imitator network with the highest MEF-SSIM score, 3) optimizing the latent code instead of the style code, 4) replacing the soft label with the hard label.
The quantitative results are presented in Table \ref{tab:ablation}.
It can be observed that the method of picking $I_{gt}^*$ prefers to pick FMMEF since it ranks first in the MEFB.
The method of picking $I_{o}^*$ performs better than picking $I_{gt}^*$,
which indicates that the imitator network improves the individuals' ability while combining them.
The method of optimizing the latent code behaves badly, as it introduces 
severe color distortion in the image. 
The method without soft label introduces artifacts that degrades the performance. 
Finally, our method of optimizing the style code performs the best,
which demonstrates the effectiveness of our optimization and random soft label. The qualitative evaluation of the compared methods is shown in Fig. \ref{fig:ablations}.

\section{Conclusion}
In this paper, we propose an ensemble-based MEF method, named EMEF. 
Extensive comparisons in the benchmark have provided evidence for 
the feasibility and effectiveness of EMEF.
The ensemble framework also has the potential to be used in other image generation tasks.

\section*{Acknowledgments}
This work was partially supported by NSFC (No. 61802322) and Natural Science Foundation of Xiamen, China (No. 3502Z20227012).

\bibliography{aaai23}

\end{document}